\documentclass[letterpaper]{article} 
\usepackage{aaai25}  
\usepackage{times}  
\usepackage{helvet}  
\usepackage{courier}  
\usepackage[hyphens]{url}  
\usepackage{graphicx} 
\usepackage{natbib}  
\usepackage{caption} 
\usepackage{algorithm}
\usepackage{algorithmic}
\usepackage{subfiles}
\usepackage{subfigure}
\usepackage{amsmath}
\usepackage{bm}
\usepackage{booktabs}
\usepackage{multicol}
\usepackage{multirow}
\usepackage{colortbl}
\usepackage{amssymb}
\usepackage{bbding}
\usepackage{pifont}
\usepackage{tikz}

\usepackage{pgfplots}
\usetikzlibrary{spy} %
\pgfplotsset{width=7cm,compat=1.14}
\usepackage{newfloat}
\usepackage{listings}
\DeclareCaptionStyle{ruled}{labelfont=normalfont,labelsep=colon,strut=off} 

\usepackage{xcolor}
\usepackage[capitalize]{cleveref}
\crefname{section}{Sec.}{Secs.}
\Crefname{section}{Section}{Sections}
\Crefname{table}{Table}{Tables}
\crefname{table}{Tab.}{Tabs.}
\def\pm{{PoseMamba}}
\definecolor{mygray}{gray}{.95}
\definecolor{mylightergray}{gray}{.99}
\definecolor{mygreen}{RGB}{10, 179, 33}

\newcommand{\thickhline}{%
\noalign {\ifnum 0=`}\fi \hrule height 1pt
\futurelet \reserved@a \@xhline
}

\floatstyle{ruled}
\newfloat{listing}{tb}{lst}{}
\floatname{listing}{Listing}
\title{PoseMamba: Monocular 3D Human Pose Estimation with Bidirectional Global-Local Spatio-Temporal State Space Model}
\author{
Yunlong Huang\textsuperscript{\rm 1},
Junshuo Liu\textsuperscript{\rm 1},
Ke Xian\textsuperscript{\rm 1,\thanks{Corresponding author}},
Robert Caiming Qiu\textsuperscript{\rm 1}
}
\affiliations{
\textsuperscript{\rm 1}Huazhong University of Science and Technology, Wuhan, China\\
\{huangyunlong, junshuo\_liu, kxian, caiming\}@hust.edu.cn
}

\usepackage{bibentry}

\begin{document}

\maketitle

\begin{abstract}

Transformers have significantly advanced the field of 3D human pose estimation (HPE). However, existing transformer-based methods primarily use self-attention mechanisms for spatio-temporal modeling, leading to a quadratic complexity, unidirectional modeling of spatio-temporal relationships, and insufficient learning of spatial-temporal correlations. Recently, the Mamba architecture, utilizing the state space model (SSM), has exhibited superior long-range modeling capabilities in a variety of vision tasks with linear complexity. In this paper, we propose PoseMamba, a novel purely SSM-based approach with linear complexity for 3D human pose estimation in monocular video. Specifically, we propose a bidirectional global-local spatio-temporal SSM block that comprehensively models human joint relations within individual frames as well as temporal correlations across frames. Within this bidirectional global-local spatio-temporal SSM block, we introduce a reordering strategy to enhance the local modeling capability of the SSM. This strategy provides a more logical geometric scanning order and integrates it with the global SSM, resulting in a combined global-local spatial scan. We have quantitatively and qualitatively evaluated our approach using two benchmark datasets: Human3.6M and MPI-INF-3DHP. Extensive experiments demonstrate that PoseMamba achieves state-of-the-art performance on both datasets while maintaining a smaller model size and reducing computational costs. The code and models will be released.

\end{abstract}

\vspace{-0.4cm}
\section{Introduction}
3D human pose estimation from monocular observations is a fundamental task in computer vision with various real-world applications~\cite{mehta2017vnect,wiederer2020traffic, czech2022board, bauer2023weakly,munea2020progress}. Typically, this involves two separate steps: 2D pose detection to locate keypoints on the image plane, followed by 2D-to-3D lifting to determine joint positions in 3D space from 2D keypoints. Recovering accurate 3D pose from 2D keypoints is challenging due to depth ambiguity and self-occlusion in monocular data. To address these challenges, significant advancements in deep learning approaches have been made, consistently improving performance~\cite{liu2020attention, chen2020anatomy, zeng2020srnet_ECCV,wang2020motion}.
\begin{figure}[htp]
\centering
\includegraphics[width=1.0\linewidth]{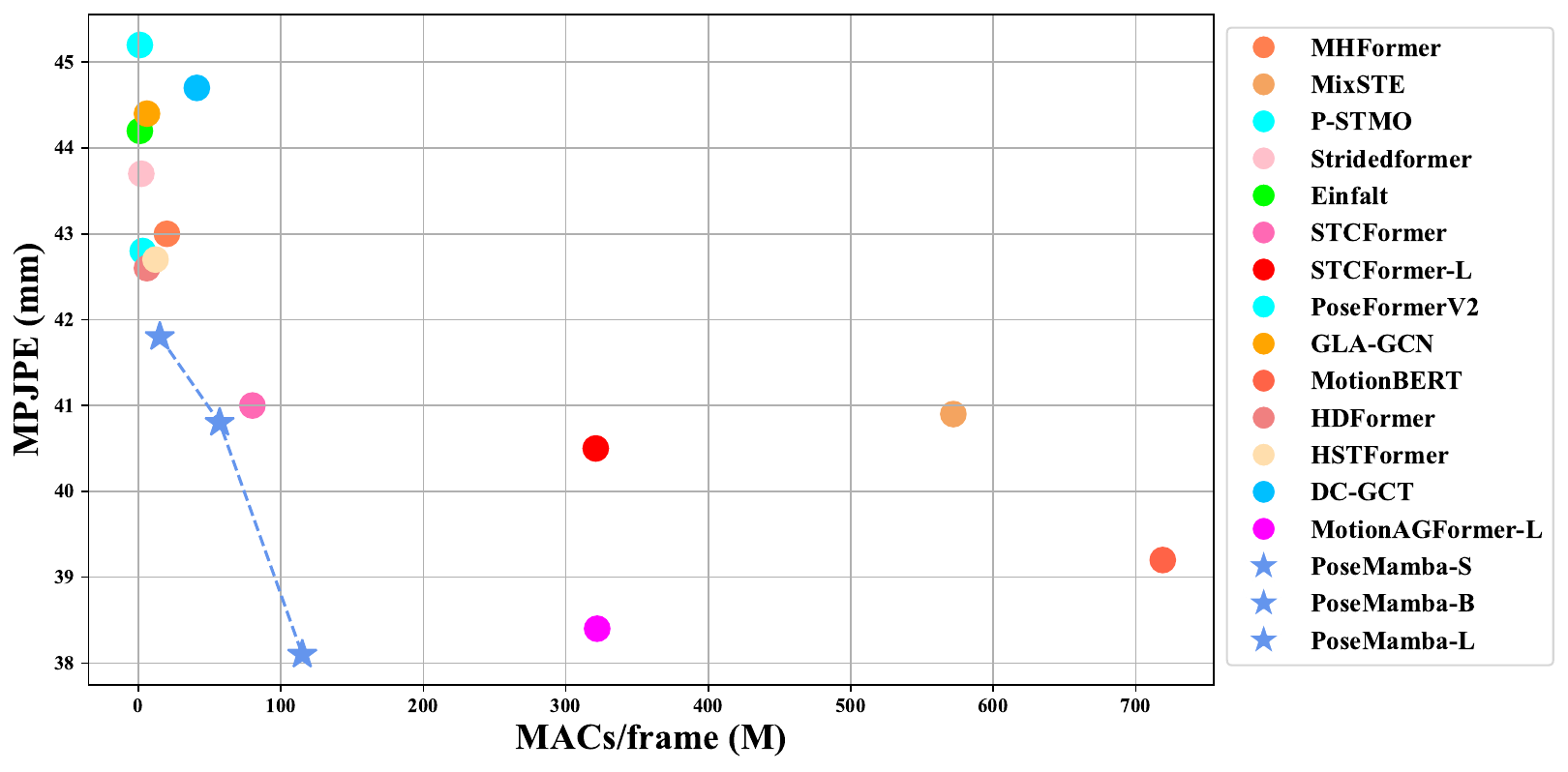}
\vspace{-0.8cm}
      \caption{Comparisons of recent 3D human pose estimation techniques on Human3.6M~\cite{ionescu2013human3} (lower is better). MACs/frame represents multiply-accumulate operations for each output frame. Our PoseMamba method presents various versions and achieves superior results, while maintaining computational efficiency.}
\label{fig:01}
		\vspace{-0.4cm}
\end{figure}
Recently, transformers~\cite{vaswani2017attention} have demonstrated significant potential in 3D human pose estimation. Its self-attention mechanism enables it to efficiently capture spatio-temporal relationships for this domain. For example, PoseFormer~\cite{poseformer} leverages spatio-temporal information to estimate more accurate central-frame pose in video sequence. MHFormer~\cite{li2022mhformer} learns spatio-temporal representations of multiple pose hypotheses in an end-to-end manner. MixSTE~\cite{mixste} proposes an alternating design using a transformer-based seq2seq model to capture the coherence between sequences. However, applying full attention mechanisms to long 2D keypoints sequence results in a notable rise in computational requirements, due to the quadratic complexity of attention calculations in both computation and memory. This naturally raises the question: \textit{how can a method be designed to function with linear complexity while still preserving the advantages of capturing spatio-temporal information?}

We observe recent progress in state space models~\cite{gu2023mamba,wang2023selective,islam2022long}, particularly with the emergence of the structured state space squence model (S4)~\cite{gu2021efficiently} as a promising architecture for sequence modeling.
Building upon S4, Mamba~\cite{gu2023mamba} incorporates time-varying parameters into the SSM, introducing an efficient hardware-aware algorithm with global receptive fields and linear complexity. Recently, a few concurrent approaches~\cite{zhu2024vision,liu2024vmamba} have focused on 2D vision tasks, such as classification and segmentation.

Driven by the successes of SSM in 2D image processing, we propose Pose State Space Model (denoted as \textbf{PoseMamba}), which features bidirectional global-local spatial-temporal modeling with linear complexity. We aim to explore the potential of SSM in 3D human pose estimation. Through pilot tests, we have observed that relying solely on Mamba~\cite{gu2023mamba} may not lead to optimal performance. We hypothesize that the issue arises from the unidirectional modeling approach of the standard SSM. To address this, we propose a \textbf{bidirectional global-local spatial-temporal modeling} approach for 3D human pose estimation. Here, global refers to spatial modeling that captures the full-body pose, while local pertains to spatial modeling focused on the limbs and their detailed movements. Specifically, within this bidirectional global-local spatio-temporal SSM block, we introduce a reordering strategy to enhance the local modeling capability of the SSM. This strategy provides a more logical geometric scanning order and integrates it with the global SSM, resulting in a combined global-local spatial scan. Experimental results on Human3.6M and MPI-INF-3DHP demonstrate the effectiveness of our method. Our PoseMamba surpasses the previous state-of-the-art (SOTA) methods while having fewer parameters and MACs, demonstrating the potential of SSM in 3D human pose estimation, as shown in \Cref{fig:01}.

In summary, the main contributions of our work are: 
\begin{itemize}
\setlength\itemsep{0em}
\item To the best of our knowledge, we are the first to introduce a novel bidirectional global-local spatio-temporal modeling approach and logical geometric scanning strategy within the Mamba framework, \pm, for 3D HPE under the category of 2D-to-3D lifting.
\item  We propose bidirectional global-local spatial-temporal modeling, enabling the PoseMamba to sufficiently learn global-local spatial-temporal information with linear complexity, exploiting the human skeleton geometry.
\item \textit{Efficiency and Flexibility:} \textbf{i)} Our PoseMamba is distinguished by its lightweight design and faster speed with fewer parameters compared to previous SOTA methods, while maintaining promising accuracy. Specifically, PoseMamba is 2.8× faster than MotionAGFormer and reduces 64.7\% GPU memory when performing batch inference to achieve 3D pose from 2D pose estimation at the frame of 243.
\textbf{ii)} To accommodate diverse needs, we provide various versions of PoseMamba, allowing users to choose a balanced option between accuracy and speed based on their specific requirements.
\item Without bells and whistles, our PoseMamba model achieves state-of-the-art results on both Human3.6M and MPI-INF-3DHP datasets.
\end{itemize}
\section{Related Work}
\subsection{3D Human Pose Estimation}

Existing 3D human pose estimation methods can be categorized through two perspectives. Firstly, these methods can be divided into two types based on the input video type: multi-view and monocular approaches. Approaches that depend on multi-view inputs~\cite{zhang2021adafuse, reddy2021tessetrack, chun2023learnable} require multiple cameras capturing different perspectives, which may pose challenges in practical applications. Secondly, these methods can be divided into direct 3D HPE methods and 2D-3D lifting methods. Direct 3D HPE methods~\cite{pavlakos2018ordinal, sun2018integral, zhou2019hemlets, huang2023simhmr} derive the spatial coordinates of joints directly from video frames without intermediary steps. In contrast, 2D-3D lifting methods first employ readily available 2D pose detectors~\cite{cpn, hrnet, newell2016stacked} before elevating 2D coordinates to 3D space~\cite{poseformerv2, motionbert, mixste}. Existing works~\cite{holmquist2023diffpose,shan2023diffusion} use multi-hypothesis approach to improve depth ambiguity in 3D HPE. DSED~\cite{liu2022explicit} addresses the self-occlusion problem in 3D HPE by explicitly reasoning about occlusion relationships in multi-person scenarios. 
While HumMUSS~\cite{mondal2024hummuss} first explores bidirectional SSM modeling for human motion understanding, our work introduces a novel bidirectional global-local spatio-temporal approach and logical geometric scanning strategy tailored for 3D HPE. PoseMagic~\cite{zhang2024pose} introduces a hybrid Mamba-GCN architecture, but its reliance on GCN for capturing local details may lead to insufficient detail for complex actions. In contrast, our PoseMamba captures local movement details more effectively and improves performance through its bidirectional global-local spatio-temporal modeling method and logical geometric scanning strategy.
		\vspace{-0.2cm}
\subsection{State Space Models}
		\vspace{-0.2cm}
Recently, Mamba~\cite{gu2023mamba} has achieved a significant breakthrough with its linear-time inference and efficient training methodology.
Building on the success of Mamba, MoE-Mamba~\cite{pioro2024moe} amalgamated Mixture of Experts with Mamba, unlocking the scalability potential of SSMs and achieving performance akin to Transformers. For vision applications, Vision Mamba~\cite{zhu2024vision} and VMamba~\cite{liu2024vmamba} employed bidirectional SSM blocks and the cross-scan module, respectively, to enhance data-dependent global visual context. However, the exploration of Mamba's potential in 3D human pose estimation remains untapped. In this paper, we do not simply apply SSM to pose estimation. We compare unidirectional scanning with bidirectional scanning and observe inaccuracies in limb recognition. Unlike Vision Mamba and VMamba, we enhance the spatial scanning method for 3D human pose estimation and propose bidirectional global-local spatial-temporal scanning to learn global-local spatial-temporal correlation sufficiently. 
		\vspace{-0.2cm}
\section{Preliminaries}
\label{sec:Preliminaries}
\subsubsection{State Space Model}
We can think of SSM as linear time-invariant (LTI) system that maps input $x(t) \in \mathbb{R}^{L}$ to output $y(t) \in \mathbb{R}^{L}$ via hidden state $h(t) \in \mathbb{C}^{N}$. It can be described as linear ordinary differential equations (ODEs):

\begin{equation}\label{eq:ssm}
\begin{split}
&\dot{h}(t) = \boldsymbol{A}h(t) + \boldsymbol{B}x(t)  \\
&y(t)  = \boldsymbol{C}h(t) + {D}x(t)
\end{split}
\end{equation}

Here, $\dot{h}(t)$ represents the time derivative of the hidden state vector $h(t)$, $\boldsymbol{A} \in \mathbb{C}^{N \times N}$, $\boldsymbol{B}, \boldsymbol{C} \in \mathbb{C}^{N}$, and $D \in \mathbb{C}^{1}$ represent the weighting parameters.

\subsubsection{Discretization of SSM}

To process discrete sequence inputs, continuous-time SSMs must be discretized, typically accomplished by solving the ODE followed by a simple discretization technique. Specifically, the analytical solution to \Cref{eq:ssm} can be represented as:

\begin{equation}\label{eq:ode_slt}
    h(t_b) = e^{\boldsymbol{A}(t_b-t_a)}(h(t_a) + \int_{t_a}^{t_b} \boldsymbol{B}(\tau) x(\tau) e^{-\boldsymbol{A}(\tau-t_a)} \,d\tau)
\end{equation}

Subsequently, through sampling with step size $\boldsymbol{\Delta}$ (i.e., $d\tau |_{t_i}^{t_{i+1}} =\Delta_i$), $h(t_b)$ can be discretized as:

\begin{equation}\label{eq:ode_dis}
h_b = e^{\boldsymbol{A}(\sum_{i=a}^{b-1} \Delta i)} \left( h_a + \sum^{b-1}_{i=a} \boldsymbol{B}_i x_i e^{-\boldsymbol{A}(\sum_{j=a}^{i} \Delta j)} \Delta_i \right)
\end{equation}

Notably, this discretization approach is roughly equivalent to the outcome achieved through the zero-order hold (ZOH) technique\cite{gu2023mamba}, commonly found in SSM-related literature.

To provide a specific example, when $b=a+1$, \Cref{eq:ode_dis} can be expressed as:

\begin{align}\label{eq:ode_dis_step}
h_{a+1} &=  \boldsymbol{\overline{A_a}} h_a + \boldsymbol{\overline{B_a}} x_a 
\end{align}

Here, $\boldsymbol{\overline{A_a}}=e^{\boldsymbol{A}\Delta_a}$ corresponds to the ZOH discretization result~\cite{gu2023mamba}, while $\boldsymbol{\overline{B_a}}=\boldsymbol{B}_a \Delta_a$ essentially represents the first-order Taylor expansion of the ZOH-derived equivalent.

\subsubsection{Selective Scan}

The weight matrix $\boldsymbol{B}$ in \Cref{eq:ode_slt} and \Cref{eq:ode_dis}, along with $\boldsymbol{C}$, $\boldsymbol{D}$, and $\boldsymbol{\Delta}$, is tailored to be input-dependent to overcome the limitations of LTI SSMs (\Cref{eq:ssm}) in capturing contextual details~\citep{gu2023mamba}. However, the introduction of time-varying SSMs presents a computational challenge because convolutions with dynamic weights are not supported, making them unsuitable for this purpose. Nonetheless, deriving the recurrence relation of $h_b$ in \Cref{eq:ode_dis} enables efficient computation. Specifically, if we define $e^{\boldsymbol{A}(\Delta_a+...+\Delta_{i-1})}$ as $\boldsymbol{p_{A, a}^{i}}$, its recurrence relation can be expressed as 

\begin{equation}\label{eq:recur_pA}
\boldsymbol{p_{A, a}^{i}} = e^{\boldsymbol{A}\Delta_{i-1}} \boldsymbol{p_{A, a}^{i-1}}
\end{equation}

Regarding the second term of \Cref{eq:ode_dis}, we obtain

\begin{align}\label{eq:recur_pB}
\boldsymbol{p_{B, a}^{b}} &= e^{\boldsymbol{A}(\Delta_a+...+\Delta_{b-1})} \sum^{b-1}_{i=a} \boldsymbol{B_i} x_i e^{-\boldsymbol{A}(\Delta_a+...+\Delta_i)} \Delta_i 
\end{align}

Therefore, utilizing the relationships derived in \Cref{eq:recur_pA} and \Cref{eq:recur_pB}, the computation of $h_{b} = \boldsymbol{p_{A, a}^{b}} h_{a} + \boldsymbol{p_{B, a}^{b}}$ can be efficiently parallelized using associative scan algorithms~\cite{martin2018parallelizing,smith2022simplified}, which are facilitated by various contemporary programming libraries. 
\section{PoseMamba}
\begin{figure}[htbp]
\centering
    \includegraphics[width=0.7\linewidth]{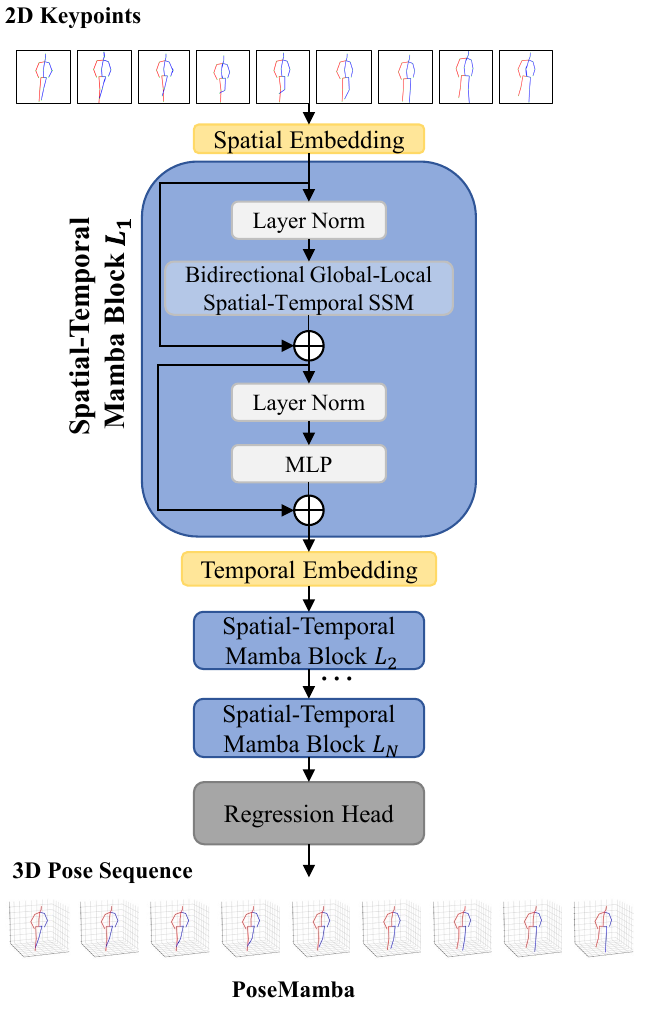}
    \vspace{-0.5cm}
\caption{The pipeline of our \pm. We start by using fully connected layer to project the input keypoint sequence, and then embed position and temporal embedding matrix into sequence. After that, we feed the sequence into the Mamba blocks.}
\label{fig:pipeline}
    \vspace{-0.5cm}
\end{figure}

As illustrated in \Cref{fig:pipeline}, our network processes a concatenated 2D coordinate array $C_{T,J}\in \mathbb{R}^{T\times J\times 2}$ representing $J$ joints across $T$ frames. The input has a channel size of 2.

Initially, we project the input keypoint sequence $C_{T,J}$ into a high-dimensional feature $P_{T,J} \in \mathbb{R}^{T \times J \times d_m}$ with each joint represented by a feature dimension of $d_m$.
Subsequently, we incorporate a spatial and a temporal position embedding matrix to preserve positional details across spatial and temporal domains.
The proposed PoseMamba takes $P_{T,J}$ as input and focuses on capturing global bidirectional spatial-temporal information efficiently through Mamba blocks with linear complexity.
Lastly, we employ a regression head to combine the encoder's outputs $Z \in \mathbb{R}^{T\times J\times d_m}$, adjusting the dimension from $d_m$ to 3 to derive the 3D human pose sequence $Out \in \mathbb{R}^{T\times J\times 3}$.
\subsection{Spatio-Temporal Encoder}
\subsubsection{Transformer-Based Spatio-Temporal Correlation Learning}
Prior transformer-based studies have primarily concentrated on utilizing multi-head self-attention mechanisms to understand spatio-temporal relationships, as illustrated in \cref{fig_attention_case}. The computation of attention for the query, key, and value matrices $Q, K, V$ in each head is expressed as:
\begin{equation}
    \begin{aligned}
        Attention(Q,K,V)=Softmax(\frac{QK^T}{\sqrt{d_m}})V,
    \end{aligned}
\end{equation}
where $\{Q,K,V\} \in \mathbb{R}^{O \times d_m}$, $O$ indicates the number of tokens, and $d_m$ is the dimension of each token.
\subsubsection{Bidirectional Global-Local Spatio-Temporal Modeling}
\begin{figure}[t]
\centering
\includegraphics[width=1\linewidth]{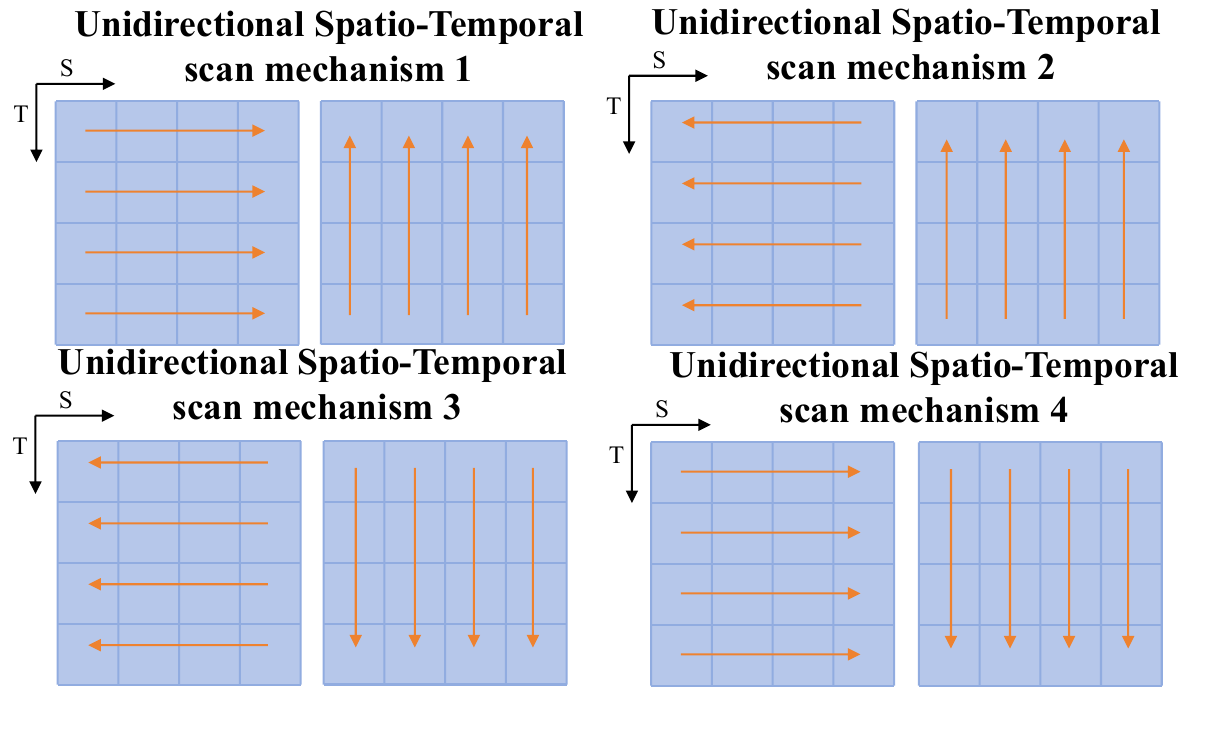}
	\vspace{-0.8cm}
\caption{Illustration of different unidirectional spatio-temporal scan mechanisms.} 
\label{fig:ubi_st_scan}
	\vspace{-0.6cm}
\end{figure}

In contrast to prior methods using attention mechanisms with quadratic computational complexity, we propose a state space model to encapsulate comprehensive spatio-temporal information at a linear complexity.
\begin{figure}[!ht]
    \vspace{-0.4cm}
\centering
\subfigure[Self-attention mechanism]
{
\label{fig_attention_case}
    \begin{minipage}[b]{0.46\linewidth}
        \centering
        \includegraphics[width=1\linewidth]{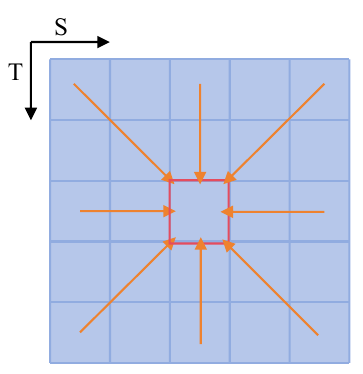}
    \end{minipage}
}
\subfigure[Bidirectional Spatio-Temporal scan mechanism]
{
    \label{fig_st_scan_case}
    \begin{minipage}[b]{0.46\linewidth}
        \centering
        \includegraphics[width=1\linewidth]{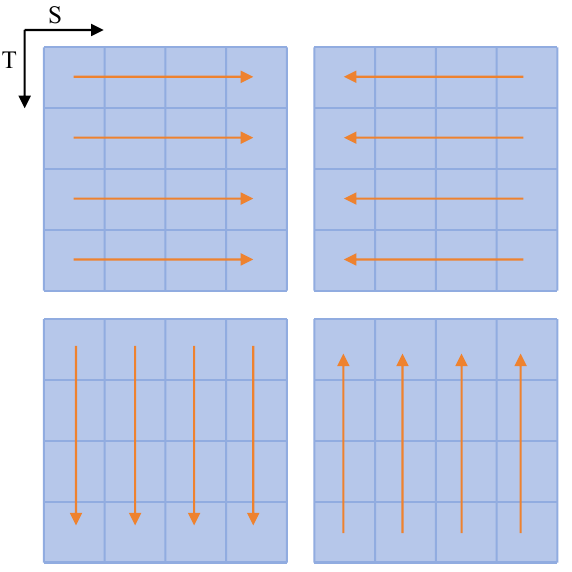}
    \end{minipage}
}
\subfigure[Bidirectional Global-Local Spatio-Temporal scan mechanism]
{
    \label{fig_global_local_st_scan_case}
    \begin{minipage}[b]{1\linewidth}
        \centering
        \includegraphics[width=1\linewidth]{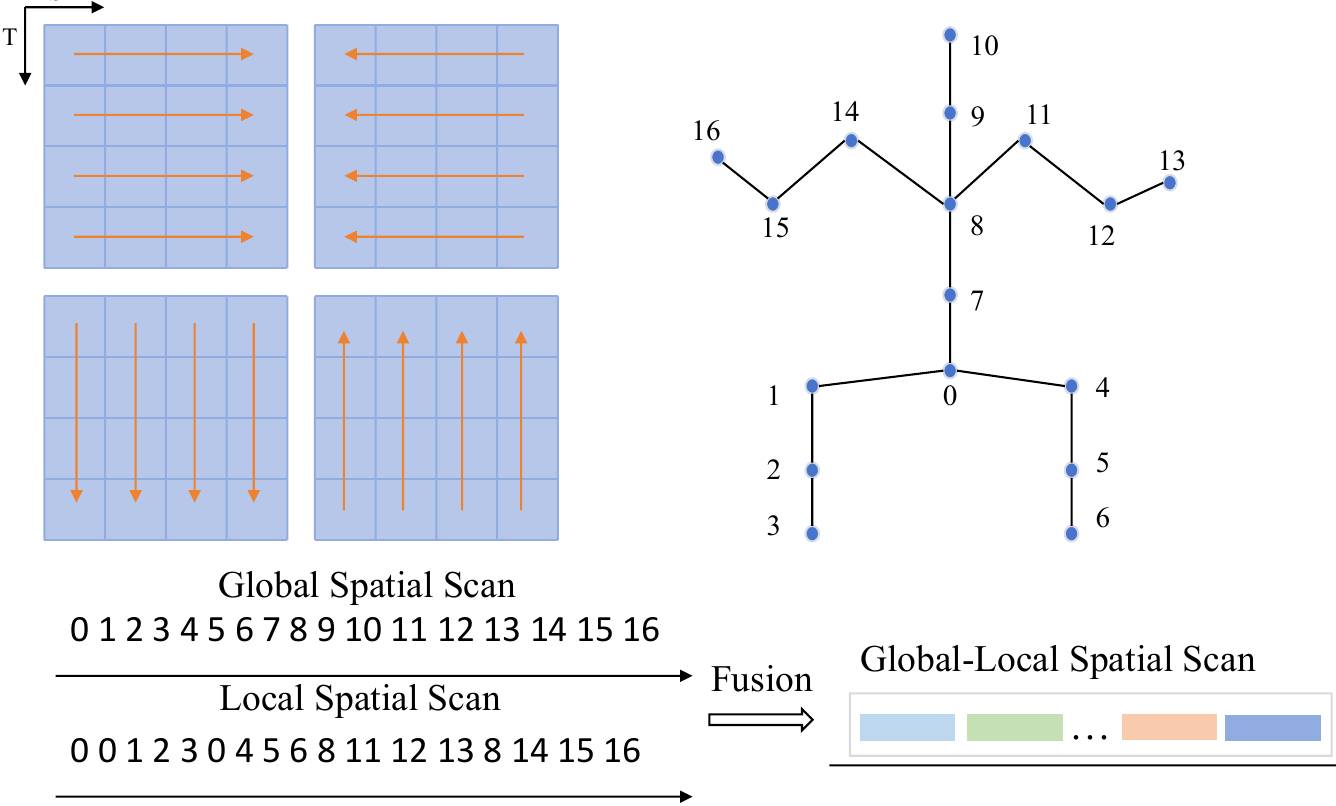}
    \end{minipage}
}
    \vspace{-0.4cm}
\caption{Illustration of various spatio-temporal modeling mechanisms. (a) Self-attention~\cite{vaswani2017attention, ViT}. (b) Bidirectional spatio-temporal scan~\cite{liu2024vmamba}. (c) Our proposed bidirectional global-local spatio-temporal scan mechanism, which leverages the geometry of the human skeleton to enhance detail.}
\label{fig:attention_compare}
    \vspace{-0.3cm}
\end{figure}
Specifically, inspired by VMamba~\cite{liu2024vmamba}, before inputting the tokens into the S6 model, we reorganize the tokens in both spatial and temporal dimensions, specifically forward spatial scan, forward temporal scan, backward spatial scan, and backward temporal scan, as depicted in \Cref{fig_st_scan_case}. Subsequently, the resultant features are merged. This approach enables the model to obtain comprehensive bidirectional global spatio-temporal information from bidirectional spatial and temporal dimensions. Furthermore, the computational complexity remains at linear complexity in contrast to the self-attention operation with quadratic complexity in transformer \Cref{fig_attention_case}. To better demonstrate the benefits of bidirectional spatio-temporal modeling, we conduct experiments on four unidirectional spatio-temporal scan mechanisms, as depicted in \Cref{fig:ubi_st_scan}, which demonstrates that relying solely on Mamba can not achieve optimal performance. 

Furthermore, to address the persistent challenge of inaccurate limb prediction, we introduce a novel reordering strategy designed to augment the local modeling capabilities of the state space model. This enhancement is achieved by establishing a more rational geometric scanning sequence, which is then seamlessly integrated with the global SSM framework. This integration facilitates a comprehensive global-local spatial scanning approach, as illustrated in \Cref{fig_global_local_st_scan_case}. Our proposed strategy not only refines the spatial scanning process but also ensures a harmonious fusion of local details with the broader spatial context, thereby significantly improving the precision of limb predictions. 
Specifically,  we posit that scanning key points on the human skeleton from 0 to 16 enables the extraction of global spatial features. However, our experimental findings indicate that relying only on global scanning consistently led to inaccurate limb prediction. Therefore, exploiting the interactions between body joints, we propose a local scanning approach to capture local human skeleton details, as detailed in \Cref{fig_global_local_st_scan_case}. We design a global-local spatial scanning approach by merging these two scanning sequences. Additionally, by incorporating temporal scanning, we develop a bidirectional global-local spatio-temporal mamba block, advancing the modeling of spatio-temporal features for 3D HPE.

\subsubsection{Bidirectional Global-Local Spatio-Temporal Mamba Block}
For each spatio-temporal Mamba block, layer normalization (LN), bidirectional spatio-temporal SSM, depth-wise convolution~\cite{chollet2017xception}, and residual connections are employed. A spatio-temporal Mamba block is shown in \Cref{fig:pipeline}, and the output can be summarized as follows:
\begin{equation}
\begin{aligned}
Z'_l &=LN(SSM(\sigma(DW(LN(Z_{l-1})))))+ Z_{l-1},\\
Z_l &=MLP(LN(Z'_l)) +Z'_l,\\
\end{aligned}
\end{equation}
where $Z_l \in \mathbb{R} ^{T\times J \times C}$ is the output of the $l$-th block. \textit{DW} means the depth-wise convolution. Following the DW, a SiLU~\cite{hendrycks2016gaussian} and SSM are adopted. 

\subsubsection{Spatio-Temporal Correlation Learning}
We employ the bidirectional global-local spatio-temporal Mamba blocks to learn spatio-temporal correlations among joints in over frames.
Firstly, we take 2D keypoints sequence as input $C_{T,J}\in \mathbb{R}^{T\times J\times 2}$ and project each keypoint to a high-dimensional feature $P_{T,J} \in \mathbb{R}^{T \times J \times d_m}$ with the linear embedding layer. 
We then embed the spatial position information with a positional matrix $E_{spos} \in \mathbb{R}^{J \times d_m}$. Each joint token $p \in P_{J}$ is projected from joint $c_i$ of the 2D coordinates $C_{J} \in \mathbb{R}^{J \times 2}$: 
\begin{equation}
    X = Norm(L_e(c_i)+E_{spos}), \ X \in \mathbb{R}^{J\times d_m},
\end{equation}
where $Norm$ denotes the layer normalization, and $L_e$ indicates the linear embedding layer.

Subsequently, the features are fed into a bidirectional spatio-temporal Mamba block to model dependencies across all joints.
We also embed the temporal position information with a temporal positional matrix $E_{tpos} \in \mathbb{R}^{T \times d_m}$:
\begin{equation}
    X = Norm(X+E_{tpos}), \ X \in \mathbb{R}^{T\times d_m},
\end{equation}
where $Norm$ denotes the layer normalization.

Then, it is fed into spatio-temporal Mamba block to model dependencies across all joints.
Finally, we obtain spatio-temporal features through $N-2$ layers of bidirectional spatio-temporal mamba blocks.
In the regression head, a linear layer is applied on the output $Z$ to perform regression to produce the 3D pose sequence $Out \in \mathbb{R}^{T\times J\times 3}$.  
	\vspace{-0.3cm}
\subsection{Loss Function}
\label{sec:multiloss}

Following the previous work~\cite{motionbert, mixste}, the network is trained in an end-to-end manner and the final loss function $\mathcal{L}$ is defined as:
\begin{equation}
    \mathcal{L} = \mathcal{L}_{3D} + \lambda_t \mathcal{L}_t + \lambda_m \mathcal{L}_m + \lambda_{2D} \mathcal{L}_{2D},
\end{equation}
where $\mathcal{L}_{3D}$ is the MPJPE loss, $\mathcal{L}_t$ is the TCLoss~\cite{hossain2018exploiting} to generate smooth poses, $\mathcal{L}_m$ denotes the MPJVE loss~\cite{pavllo20193d} to improve the temporal coherence, and $\mathcal{L}_{2D}$ denotes the 2D re-projection loss~\cite{motionbert}. 
During the training stage, different coefficients $\lambda_t$ and $\lambda_m$ are employed to $\mathcal{L}_t$ and $\mathcal{L}_m$ to avoid excessive smoothness in sequence. We merge the TCLoss and MPJVE as the temporal loss function (T-Loss) inspired by the previous work~\cite{mixste}. 
The MPJPE loss $L_{3D}$ is computed as follows:
\begin{equation}
    \mathcal{L}_{3D}=\sum_{t=1}^{T} \sum_{i=1}^{J}\left\|Y_{i}^{t}-\widetilde{X}_{i}^{t}\right\|_{2},
\end{equation}
where $\widetilde{X}_{i}^{t}$ and $Y_{i}^{t}$ represent the predicted and ground truth 3D poses of joint $i$ at frame $t$, respectively. 
\begin{table*}[!ht]\small
\caption{Quantitative comparisons on Human3.6M.
$T$: Number of input frames. CE: Estimating center frame only. P1: MPJPE error (mm). P2: P-MPJPE error (mm). ${\mathrm{P1}^\dag}$: P1 error on 2D ground truth. (*) denotes using HRNet~\cite{hrnet} for 2D pose estimation. The best and second-best scores are in bold and underlined, respectively.}
  \centering
    \vspace{-0.3cm}
        \resizebox{1\linewidth}{!}{
    \begin{tabular}{lcc|ccccc}
      \hline
      Method & $T$ & CE & Param & MACs & MACs/frame & P1$\downarrow$	
/P2$\downarrow$	 & ${\mathrm{P1}^\dag}$$\downarrow$	 
 \\
      \hline
      *MHFormer~\cite{li2022mhformer} CVPR'22 & 351 & \checkmark & 30.9 M & 7.0 G & 20 M & 43.0/34.4 & 30.5 \\
      MixSTE~\cite{mixste} CVPR'22 & 243 & $\times$ & 33.6 M & 139.0 G & 572 M & 40.9/32.6 & 21.6\\
      P-STMO~\cite{shan2022p} ECCV'22 & 243 & \checkmark & 6.2 M & 0.7 G & 3 M & 42.8/34.4 & 29.3\\
      Stridedformer~\cite{li2022exploiting} TMM'22 & 351 & \checkmark & 4.0 M & 0.8 G &2 M &43.7/35.2 & 28.5\\
      Einfalt~\textit{et al.}~\cite{einfalt_up3dhpe_WACV23} WACV'23 & 351 & \checkmark & 10.4 M & 0.5 G & 1 M & 44.2/35.7 & - \\
      STCFormer~\cite{STCFormer} CVPR'23  & 243 & $\times$ & 4.7 M & 19.6 G & 80 M & 41.0/32.0 & 21.3 \\
      STCFormer-L~\cite{STCFormer} CVPR'23 & 243 & $\times$ & 18.9 M & 78.2 G & 321 M & 40.5/\underline{31.8} & - \\
      PoseFormerV2~\cite{poseformerv2} CVPR'23 & 243 & \checkmark & 14.4 M & 4.8 G & 20 M & 45.2/35.6 & -\\
      GLA-GCN~\cite{yu2023gla} ICCV'23 & 243 & \checkmark & 1.3 M & 1.5 G & 6 M & 44.4/34.8 & 21.0\\
      MotionBERT~\cite{motionbert} ICCV'23 & 243 & $\times$ & 42.3 M & 174.8 G & 719 M & 39.2/32.9 & 17.8\\
      HDFormer~\cite{chen2023hdformer} IJCAI'23 & 96 & $\times$ & 3.7 M & 0.6 G & 6 M & 42.6/33.1 & 21.6\\
      HSTFormer~\cite{qian2023hstformer} arXiv'23 & 81 & $\times$ & 22.7 M & 1.0 G & 12 M & 42.7/33.7 & 27.8\\
      DC-GCT~\cite{kang2023double} arXiv'23 & 81 & \checkmark & 3.1 M & 41 M & 41 M & 44.7/- & -\\
      MotionAGFormer-L~\cite{mehraban2024motionagformer} WACV'24& 243 & $\times$ & 19.0 M & 78.3 G & 322 M & 38.4/32.5 & 17.3 \\
      \rowcolor{mygray} PoseMamba-S & 243 & $\times$ & 0.9 M& 3.6 G& 15 M & 41.8/35.0 & 20.0 \\
      \rowcolor{mygray} PoseMamba-B & 243 & $\times$ & 3.4 M &13.9 G& 57 M & 40.8/34.3 & 16.8 \\
      \rowcolor{mygray} PoseMamba-L & 243 & $\times$ & 6.7 M & 27.9 G & 115 M & \underline{38.1}/32.5 & \underline{15.6} \\
      \rowcolor{mygray} PoseMamba-X & 243 & $\times$ & 26.5 M & 109.9 G & 452 M & \textbf{37.1}/\textbf{31.5} & \textbf{14.8} \\
      \hline
    \end{tabular}}
\label{tab:human3.6m-comparison}
    \vspace{-0.5cm}
\end{table*}
	\vspace{-0.3cm}
\section{Experiment}
We evaluate our proposed PoseMamba on two large-scale 3D human pose estimation datasets, i.e., Human3.6M~\cite{ionescu2013human3} and MPI-INF-3DHP~\cite{mehta2017monocular}.
\subsection{Datasets and Evaluation Metrics}

\textbf{Human3.6M} is a commonly used indoor dataset for 3D human pose estimation. It contains 3.6 million video frames of 11 subjects performing 15 different daily activities. To ensure fair evaluation, we follow the standard approach and train the model using data from subjects 1, 5, 6, 7, and 8, and then test it on data from subjects 9 and 11. Following the previous work~\cite{motionbert}, we use two protocols for evaluation. The first protocol (referred to as P1) uses Mean Per Joint Position Error (MPJPE) in millimeters between the estimated pose and the actual pose, after aligning their root joints (sacrum). The second protocol (referred to as P2) measures Procrustes-MPJPE, where the actual pose and the estimated pose are aligned through a rigid transformation.
\textbf{MPI-INF-3DHP} is another large-scale dataset gathered in three different settings: green screen, non-green screen, and outdoor environments. This dataset has 1.3 million frames, containing a wider range of movements than Human3.6M. We utilize MPJPE as the evaluation metric.
\subsection{Implementation Details}
\subsubsection{Model Variants} We create three model configurations, detailed in Table~\ref{tab:variants}. Our base model, \pm-B, balances accuracy and computational cost. Other variants are named based on parameters and computational needs. The selection of each variant depends on specific application needs, like real-time processing or precise estimations. The MLP's expansion layer is $\alpha = 2$ for all experiments.
\begin{table}[ht]\small
    \caption{PoseMamba model variants. $N$: Number of layers. $d_m$:  Dimension of model. $T$: Number of input frames.}
\vspace{-0.3cm}
        \begin{tabular}{lccccc}
            \toprule
            Method & $N$ & $d_m$ & $T$ & Params & MACs \\
            \midrule
            \pm-S & 20 & 64 & 243 & 0.860 M & 3.587 G\\
            \pm-B & 20 & 128 & 243 & 3.358 M & 13.943 G\\
            \pm-L & 40 & 128 & 243 & 6.714 M & 27.881 G\\
            \pm-X & 40 & 256 & 243 & 26.535 M & 109.909 G\\
            \bottomrule
        \end{tabular}%
    \label{tab:variants}
    \vspace{-0.3cm}
\end{table}
\begin{table}[!ht]\small
\caption{Quantitative comparisons on MPI-INF-3DHP. $T$: Number of input frames. The best and second-best scores are in bold and underlined, respectively.}
  \centering
\vspace{-0.3cm}
  \resizebox{1\linewidth}{!}{%
    \begin{tabular}{lc|ccc}
      \hline
      Method & $T$ & MPJPE$\downarrow$ \\
      \hline
      MHFormer~\cite{li2022mhformer} & 9 & 58.0 \\
      MixSTE~\cite{mixste} & 27  & 54.9 \\
      P-STMO~\cite{shan2022p} & 81  & 32.2 \\ 
      Einfalt \textit{et al.}~\cite{einfalt_up3dhpe_WACV23} & 81  & 46.9 \\
      STCFormer~\cite{STCFormer} & 81  & 23.1 \\
      PoseFormerV2~\cite{poseformerv2} & 81 & 27.8 \\
      GLA-GCN~\cite{yu2023gla} & 81 & 27.7 \\
      HSTFormer~\cite{qian2023hstformer} & 81 & 41.4 \\
      HDFormer~\cite{chen2023hdformer} & 96  & 37.2 \\
       MotionAGFormer-XS & 27  & 19.2 \\
       MotionAGFormer-S & 81 & 17.1 \\
       MotionAGFormer-B & 81  & 18.2 \\
       MotionAGFormer-L & 81  & 16.2 \\
      \rowcolor{gray!10} \pm-S & 27 &  17.79 \\
      \rowcolor{gray!10} \pm-S & 81 &  \underline{15.27} \\
      \rowcolor{gray!10} \pm-B & 81 &  \textbf{14.51} \\
      \hline
    \end{tabular}%
  }
\label{tab:MPI-INF-3DHP-comparison}
	\vspace{-0.6cm}
\end{table}
 \subsubsection{Experimental settings} 
Our model is developed utilizing PyTorch and deployed on one NVIDIA RTX 3090 GPU. Horizontal flipping augmentation is applied for both training and testing, as outlined in ~\cite{motionbert, poseformerv2}. During model training, the batch size is configured with 4 sequences. The optimization of network parameters is carried out using the AdamW~\cite{AdamW} optimizer across 120 epochs with a weight decay of 0.01. The initial learning rate is established at $2e^{-4}$ with an exponential learning rate decay schedule, utilizing a decay factor of 0.99. In our approach, we leverage the Stacked Hourglass~\cite{newell2016stacked} 2D pose detection outcomes and 2D ground truths sourced from the Human3.6M and MPI-INF-3DHP datasets, following MotionBERT~\cite{motionbert}. In MPI-INF-3DHP, we employ ground truth 2D detection using a methodology following methods ~\cite{poseformerv2, STCFormer}.
\subsection{Performance comparison on Human3.6M}
We present a comparative analysis of our PoseMamba model against other models using the Human3.6M dataset. To ensure a fair assessment, only the outcomes of models without additional pre-training on supplementary data are considered. The results, as detailed in Table~\ref{tab:human3.6m-comparison}, reveal that \pm-L achieves a P1 error of 38.1\,mm for estimated 2D pose and 15.6\,mm for ground truth 2D pose. Notably, these results are accomplished with only 16\% of the computational resources in comparison to the previous SOTA model, MotionBERT, while exhibiting an enhanced accuracy of 1.1\,mm and 2.2\,mm, respectively. Furthermore, our model achieves these results using only 36\% of the computational resource compared to another previous SOTA model, MotionAGFormer~\cite{mehraban2024motionagformer}, while being 0.3\,mm and 1.7\,mm more accurate, respectively. 

\subsection{Performance comparison on MPI-INF-3DHP}
When assessing our approach to the MPI-INF-3DHP dataset, we adapted our small and base models to accommodate 27 and 81 frames to suit the shorter video sequences. Our method demonstrates superior performance across all model variants compared to others in terms of MPJPE, as illustrated in Table~\ref{tab:MPI-INF-3DHP-comparison}, showcasing the excellence of our model.

\subsection{Ablation Studies}
\label{sec:ablation}

To evaluate the impact and performance of each component in our model, we evaluate their effectiveness in this section.
\subsubsection{Bidirectional Global-Local Spatio-Temporal Modeling}
\label{sec:ablation_bi_st_scan}
We perform comprehensive experiments to verify the effectiveness of modifying the crucial bidirectional global-local spatio-temporal modeling in PoseMamba on Human3.6M using our small variant version, where feature dimensions are altered to ensure comparable architectural parameters and MACs for a fair evaluation.
As shown in Table~\ref{tab:ablation_scan}, employing unidirectional spatio-temporal modeling results in a model performance of MPJPE ranging from 43.0 to 43.8\,mm, which is comparatively less efficient than the bidirectional spatio-temporal modeling yielding an MPJPE of 42.4\,mm. Furthermore, integrated with the local spatial scan to enhance accurate limb prediction, our final model is 0.6\,mm better than bidirectional spatial-temporal modeling, which indicates the efficacy of our global-local modeling.
\begin{table}[!ht]
\centering
	\vspace{-0.3cm}
    \caption{Ablation study for various spatial-temporal modeling with MPJPE on Human3.6M.}
	\vspace{-0.3cm}
    \resizebox{1\linewidth}{!}{%
     \begin{tabular}{l  |ccc| c  }
    \toprule
    Spatial-Temporal Modeling Strategy  &T&Params & MACs& MPJPE \\
    \midrule
    Unidirectional Spatial-Temporal 1& 243 & 0.860 M & 3.587 G& 43.1  \\ 
    Unidirectional Spatial-Temporal 2& 243 & 0.860 M & 3.587 G& 43.2  \\ 
    Unidirectional Spatial-Temporal 3& 243 & 0.860 M & 3.587 G& 43.8  \\ 
    Unidirectional Spatial-Temporal 4& 243 & 0.860 M & 3.587 G& 43.0 \\ 
    Bidirectional Spatial-Temporal & 243 & 0.860 M & 3.587 G& 42.4 \\
    \rowcolor[HTML]{DADADA}
    Bidirectional Global-Local Spatial-Temporal & 243 & 0.860 M & 3.587 G& \textbf{41.8} \\
    \bottomrule
    \end{tabular}}
    \label{tab:ablation_scan}
\vspace{-0.4cm}
\end{table}
\subsubsection{Effect of Loss Function}	
We explore the contribution of our loss function using our small variant version in detail.
As shown in \Cref{tab:ablation_loss}, the MPJPE metric decreases from 43.7 to 43.5\,mm after applying the 2D loss and decreases from 43.5 to 42.1\,mm after applying the T-Loss.
The result demonstrates that the T-Loss and 2D loss is an essential loss to improve accuracy. 
Finally, after applying the T-Loss, 2D-loss, and MPJPE loss to our method, the result achieves the best on the MPJPE metrics 41.8\,mm.
The results demonstrate that our loss function is comprehensive for the proposed model regarding accuracy and smoothness.
\begin{table}[ht]
    \centering
	\vspace{-0.3cm}
    \caption{Ablation study for loss function with MPJPE and PMPJPE on Human3.6M.}
	\vspace{-0.3cm}
    \resizebox{190pt}{!}{
        \begin{tabular}{l|c}
            \toprule
            Loss& MPJPE$\downarrow$	
/PMPJPE$\downarrow$ \\
            \midrule
            MPJPE Loss                         & 43.7/36.5		\\
            MPJPE Loss + 2D-Loss                         & 43.5/36.2		\\
            MPJPE Loss + T-Loss		   & 42.1/35.1	\\
            \rowcolor[HTML]{DADADA}
            Ours (MPJPE Loss + T-Loss + 2D-Loss) 	   & \textbf{41.8}/\textbf{35.0}	\\
            \bottomrule
        \end{tabular}
    }
    \label{tab:ablation_loss}
\vspace{-0.4cm}
\end{table}
\begin{table}[!ht]\small
\caption{The P1 error comparison by varying number of PoseMamba blocks and number of channels on Human3.6M. $d_m$: Number of channels in each \pm~block.  $T$ is kept 243 in all experiments.}
  \centering
    \vspace{-0.3cm}
    \begin{tabular}{cccc|c}
   \toprule
      Depth & $d_m$ & Param & MACs & P1 \\
   \midrule
      12 &64 & 0.516 M & 2.2 G  & 43.0 \\
      16 &64 & 0.688 M & 2.9 G  & 42.0 \\
      20 &64 & 0.860 M & 3.6 G  & 41.8 \\
      24 &64 & 1.031 M & 4.3 G  & 41.5 \\
      32 &64 & 1.375 M & 5.7 G  & 41.5 \\
      40 &64 & 1.719 M & 7.2 G& 41.1 \\
      48 &64 & 2.062 M & 8.6 G  & 41.1 \\
   \hline
        40      & 32              & 0.450 M & 1.9 G  & 43.1       \\
        40      & 64               & 1.719 M & 7.2 G  & 41.1      \\
        40      & 128            & 6.714 M & 27.9 G  &      38.1 \\
        40      & 256 		      & 26.535 M & 109.9 G  & 37.1      \\
        \hline
      12 &256 & 7.963 M & 33.0 G  & 39.1 \\
      20 &128  & 3.358 M &13.9 G  & 40.8 \\
      \hline
    \end{tabular}
\label{tab:ablation_params}
\vspace{-0.5cm}
\end{table}
\subsubsection{Parameter Setting Analysis}
\Cref{tab:ablation_params} shows how the setting of different hyper-parameters in our method impacts the performance under Protocol 1 with MPJPE. 
There are three main hyper-parameters for the network: the depth of PoseMamba ($N$), the dimension of model ($d_m$), and the input sequence length ($T$).
We divide the configurations into 2 groups row-wise, and different values are assigned for one hyper-parameters while keeping the other two hyper-parameters fixed to evaluate the impact and choice of each configuration.
In addition to these two sets of experiments, we have also conducted additional hyperparameter experiments.
Based on the results in the table, considering performance and efficiency, we choose 
three variants in \Cref{tab:variants}.

	\vspace{-0.3cm}
\subsection{Qualitative Analysis}
\begin{figure}[!ht]
    \centering
    \includegraphics[width=1.0\linewidth]{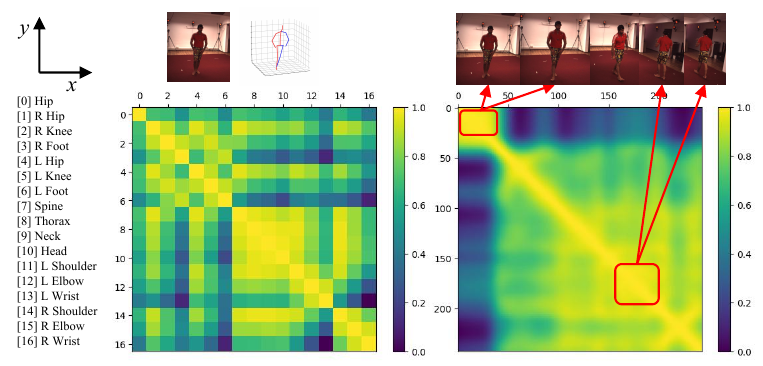}
	\vspace{-0.8cm}
    \caption{
        Visualization of SSM map among body joints and frames. 
    }
    \label{fig:attnmap}
    \vspace{-0.5cm}
\end{figure}

\begin{figure}[!ht]
\centering
\includegraphics[width=1\linewidth]{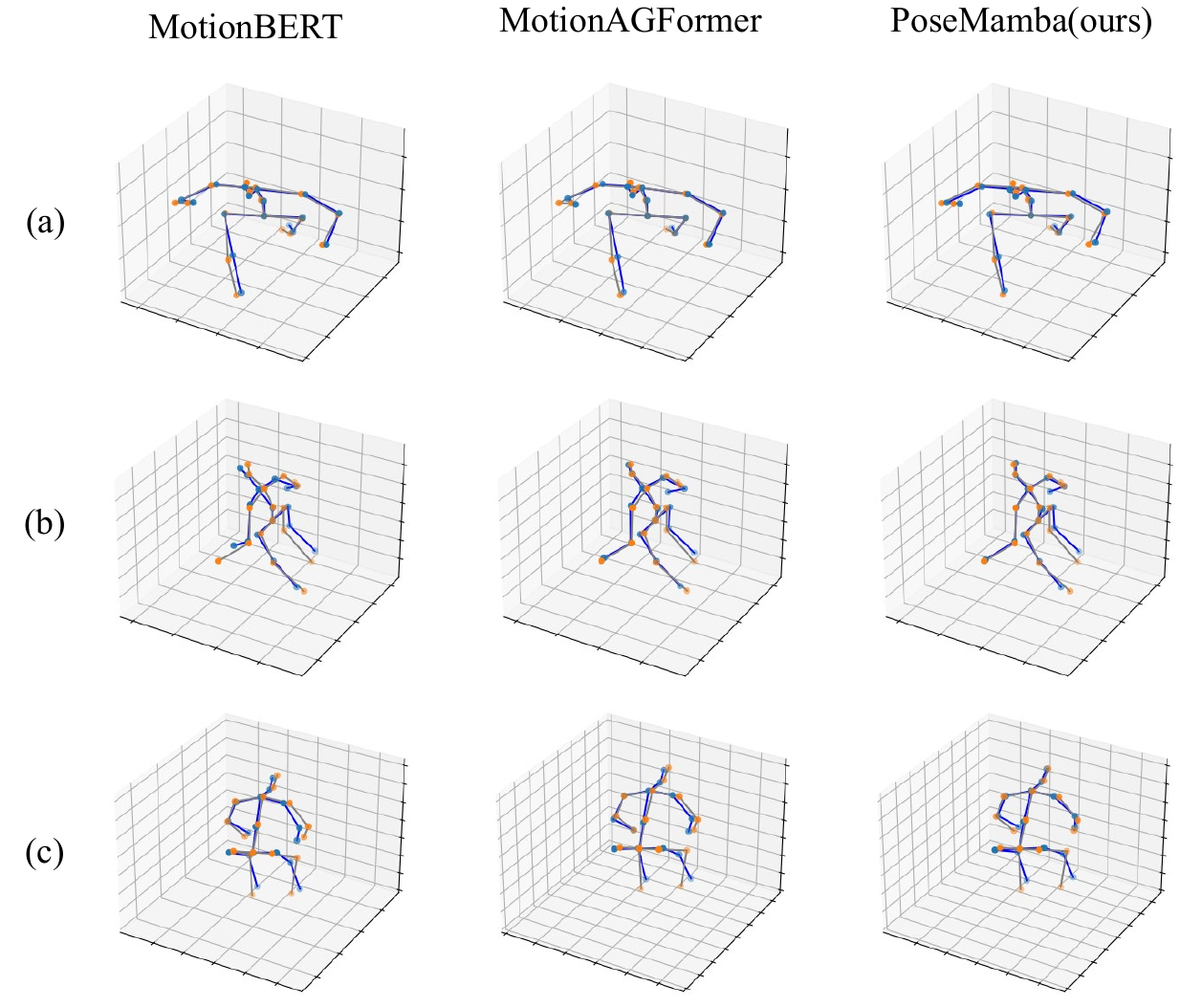}
	\vspace{-0.8cm}
\caption{Qualitative comparisons with MotionBERT and MotionAGFormer. The gray skeleton is the ground-truth 3D pose and the blue skeleton is the estimated body. }
\label{fig:qualitative-comparison}
	\vspace{-0.3cm}
\end{figure}

\begin{figure}[!ht]
\centering
\includegraphics[width=1\linewidth]{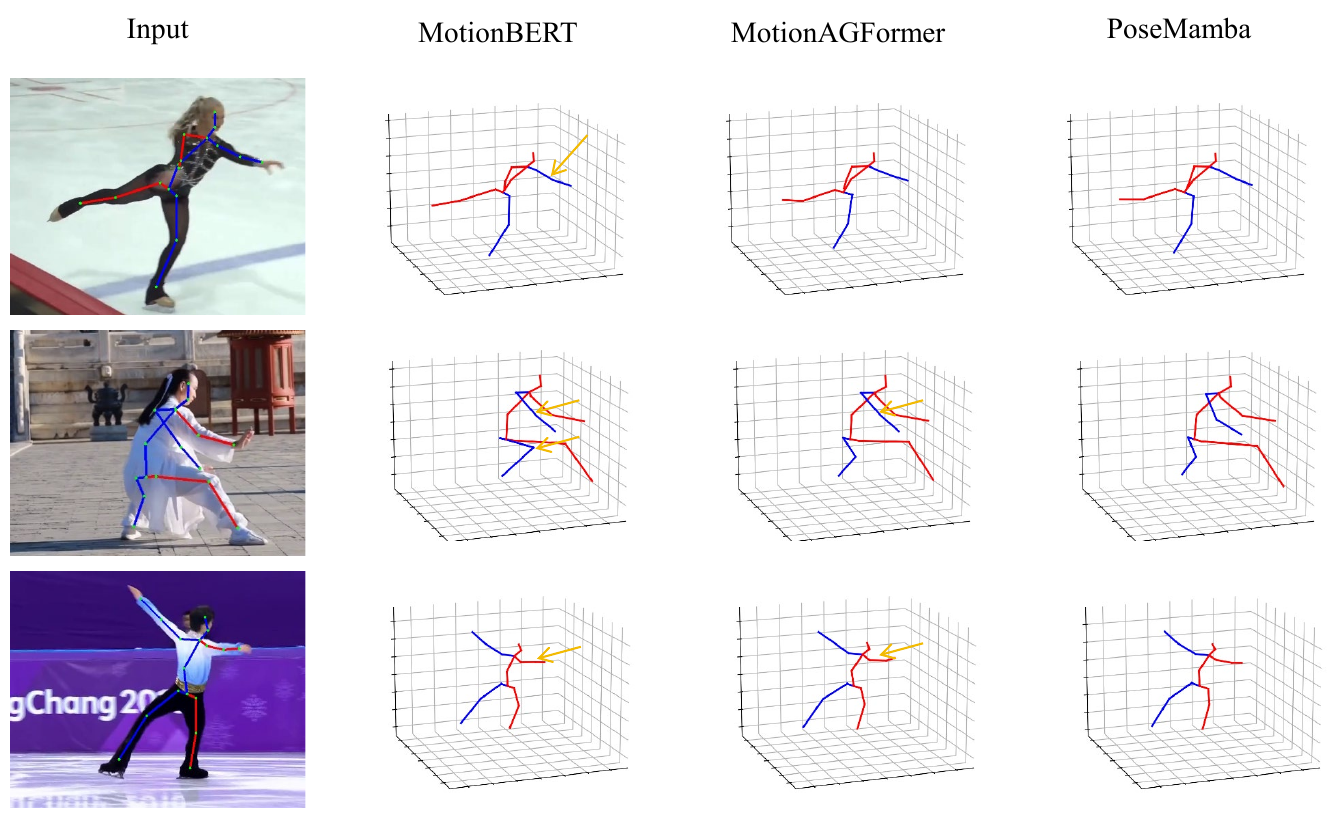}
	\vspace{-0.8cm}
\caption{Qualitative comparisons with MotionBERT and MotionAGFormer on challenging wild videos. Wrong estimations are highlighted by yellow arrows. }
\label{fig:qualitative-comparison_in_the_wild}
\vspace{-0.3cm}
\end{figure}

\Cref{fig:attnmap} visualizes last spatio-temporal SSM block map of action (\textit{Walking} of testset \textit{S9}). It can be easily observed from spatial map (left of \Cref{fig:attnmap}) that our model learns distinct dependencies between joints. Furthermore, we also visualize the temporal map (right of \Cref{fig:attnmap}). The two light-colored parts have similar poses in adjacent frames, while dark-colored frame (the middle image in the frame sequence) has a more distinct pose in adjacent frames.
~\Cref{fig:qualitative-comparison} compares \pm-L with recent approaches, which shows that our PoseMamba achieves more accurate poses than MotionBERT and MotionAGFormer. 
Moreover, Figure~\ref{fig:qualitative-comparison_in_the_wild} shows the qualitative comparison on some wild videos. It is evident that our method can produce more accurate 3D poses, particularly in cases the human action is complex and rare. 



\section{Conclusion}
We present PoseMamba, a novel SSM-based approach for 3D human pose estimation, which has a bidirectional global-local spatio-temporal mamba block to comprehensively model the human joint relations within each frame as well as the temporal correlations across frames. In the bidirectional global-local spatio-temporal mamba block, we propose a reordering strategy to enhance SSM’s local modeling ability by providing a more logical geometric scanning order and fusing it with global SSM to get global-local spatial scan. Experimental results demonstrate that PoseMamba outperforms the existing counterparts on both datasets while significantly reducing parameters and MACs. As a newcomer to 3D human pose estimation, PoseMamba is a promising option for constructing 3D vision foundation models, and we hope it can offer a new perspective for the field.

\textbf{Acknowledgements.} This work was supported by the National Natural Science Foundation of China under Grant 62406120 and the Guangxi Science and Technology Project (GuiKe-AB21196034).
\bibliography{aaai25}


\end{document}


\twocolumn[
\begin{@twocolumnfalse}
\begin{center}
\Large\textbf{PoseMamba: Monocular 3D Human Pose Estimation with Bidirectional Global-Local Spatio-Temporal State Space Model} \\
\vspace*{10pt}
\Large\textbf{Supplementary Material} \\
\vskip .5em
\end{center}
\end{@twocolumnfalse}]
This supplementary material contains the following details:
(1) Additional quantitative results. 
(2) Additional visualization results. 
\section{Additional quantitative results}
\subsection{Per-action Error Comparison}
\begin{table*}[ht]\small
\caption{Quantitative comparisons of 3D human pose estimation per action on Human3.6M. (Top) MPJPE~(mm) using detected 2D pose sequence. (Bottom) P-MPJPE~(mm) using detected 2D pose sequence. (*) denotes using HRNet~\cite{hrnet} for 2D pose estimation. (\dag) denotes manually evaluated using their provided evaluation code.}
  \centering
  \resizebox{\linewidth}{!}{
    \begin{tabular}{lc|ccccccccccccccc|c}
      \hline
      \textbf{MPJPE} & $T$ & Dire. & Disc. & Eat & Greet & Phone & Photo & Pose & Purch. & Sit & SitD & Smoke & Wait & WalkD & Walk & WalkT & Avg\\
      \hline
      *MHFormer~\cite{li2022mhformer} & 351 & 39.2 & 43.1 & 40.1 & 40.9 & 44.9 & 51.2 & 40.6 & 41.3 & 53.5 & 60.3 & 43.7 & 41.1 & 43.8 & 29.8 & 30.6 & 43.0\\
      MixSTE~\cite{mixste} & 243 & 37.6 & 40.9 & 37.3 & 39.7 & 42.3 & 49.9 & 40.1 & 39.8 & 51.7 & 55.0 & 42.1 & 39.8 & 41.0 & 27.9 & 27.9 & 40.9\\
        P-STMO~\cite{shan2022p} & 243 & 38.9 & 42.7 & 40.4 & 41.1 & 45.6 & 49.7 & 40.9 & 39.9 & 55.5 & 59.4 & 44.9 & 42.2 & 42.7 & 29.4 & 29.4 & 42.8\\
      StridedFormer~\cite{li2022exploiting} & 351 & 40.3 & 43.3 & 40.2 & 42.3 & 45.6 & 52.3 & 41.8 & 40.5 & 55.9 & 60.6 & 44.2 & 43.0 & 44.2 & 30.0 & 30.2 & 43.7\\
      Einfalt~\textit{et al.}~\cite{einfalt_up3dhpe_WACV23} & 351 & 39.6 & 43.8 & 40.2 & 42.4 & 46.5 & 53.9 & 42.3 & 42.5 & 55.7 & 62.3 & 45.1 & 43.0 & 44.7 & 30.1 & 30.8 & 44.2\\
      STCFormer~\cite{STCFormer} & 243 & 39.6 & 41.6 & 37.4 & 38.8 & 43.1 & 51.1 & 39.1 & 39.7 & 51.4 & 57.4 & 41.8 & 38.5 & 40.7 & 27.1 & 28.6 & 41.0\\
      STCFormer-L~\cite{STCFormer} & 243 & 38.4 & 41.2 & \underline{36.8} & 38.0 & 42.7 & 50.5 & 38.7 & 38.2 & 52.5 & 56.8 & 41.8 & 38.4 & 40.2 & \textbf{26.2} & 27.7 & 40.5\\
      UPS~\cite{foo2023unified} & 243 & 37.5 & 39.2 & 36.9 & 40.6 & \textbf{39.3} & \textbf{46.8} & 39.0 & 41.7 & 50.6 & 63.5 & 40.4 & 37.8 & 44.2 & 26.7 & 29.1 & 40.8\\
      \dag~MotionBERT~\cite{motionbert} & 243 & \underline{36.6} & 39.3 & 37.8 & \underline{33.5} & 41.4 & 49.9 & \underline{37.0} & 35.5 & 50.4 & 56.5 & 41.4 & 38.2 & 37.3 & \textbf{26.2} & \textbf{26.9} & 39.2\\
      HDFormer~\cite{chen2023hdformer} & 96 & 38.1 & 43.1 & 39.3 & 39.4 & 44.3 & 49.1 & 41.3 & 40.8 & 53.1 & 62.1 & 43.3 & 41.8 & 43.1 & 31.0 & 29.7 & 42.6\\
      HSTFormer~\cite{qian2023hstformer} & 81 & 39.5 & 42.0 & 39.9 & 40.8 & 44.4 & 50.9 & 40.9 & 41.3 & 54.7 & 58.8 & 43.6 & 40.7 & 43.4 & 30.1 & 30.4 & 42.7\\
      MotionAGFormer-L & 243 & 36.8 & \underline{38.5} & \textbf{35.9} & \textbf{33.0} & 41.1 & 48.6 & 38.0 & \underline{34.8} & \underline{49.0} & \textbf{51.4} & \underline{40.3} & \underline{37.4} & \textbf{36.3} & 27.2 & \underline{27.2} & \underline{38.4}\\
      \rowcolor{gray!10} PoseMamba-S & 243 & 39.5 & 41.6 & 39.9 & 35.7 & 43.5 & 52.7 & 40.2 & 37.9 & 53.1 & 63.4 & 42.4 & 40.0 & 39.6 & 29.3 & 29.2   & 41.8\\
      \rowcolor{gray!10} PoseMamba-B & 243 & 38.8 & 40.8 & 38.8 & 35.2 & 42.1 & 50.8 & 38.8 & 36.4 & 51.8 & 61.9 & 42.0 & 38.4 & 38.7 & 28.1 & 28.7&40.8\\
      \rowcolor{gray!10} PoseMamba-L & 243 & \textbf{36.5} & \textbf{37.9} & 37.5 & \underline{33.5} & \underline{39.6} & \underline{47.7} & \textbf{36.5} & \textbf{34.0} & \textbf{47.8} & \underline{54.5} & \textbf{40.2} & \textbf{36.3} & \underline{36.7} & \underline{26.4} & \textbf{26.9} & \textbf{38.1}\\
      \hline
      \hline
      \textbf{P-MPJPE} & $T$ & Dire. & Disc. & Eat & Greet & Phone & Photo & Pose & Purch. & Sit & SitD & Smoke & Wait & WalkD & Walk & WalkT & Avg\\
      *MHFormer~\cite{li2022mhformer} & 351 & 31.5 & 34.9 & 32.8 & 33.6 & 35.3 & 39.6 & 32.0 & 32.2 & 43.5 & 48.7 & 36.4 & 32.6 & 34.3 & 23.9 & 25.1 & 34.4\\
      MixSTE~\cite{mixste} & 243 & 30.8 & 33.1 & \textbf{30.3} & 31.8 & 33.1 & 39.1 & 31.1 & 30.5 & 42.5 & \textbf{44.5} & 34.0 & 30.8 & 32.7 & 22.1 & 22.9 & 32.6\\
                P-STMO~\cite{shan2022p} & 243 & 31.3 & 35.2 & 32.9 & 33.9 & 35.4 & 39.3 & 32.5 & 31.5 & 44.6 & 48.2 & 36.3 & 32.9 & 34.4 & 23.8 & 23.9 & 34.4\\
      StridedFormer~\cite{li2022exploiting} & 351 & 32.7 & 35.5 & 32.5 & 35.4 & 35.9 & 41.6 & 33.0 & 31.9 & 45.1 & 50.1 & 36.3 & 33.5 & 35.1 & 23.9 & 25.0 & 35.2\\
      Einfalt~\textit{et al.}~\cite{einfalt_up3dhpe_WACV23} & 351 & 32.7 & 36.1 & 33.4 & 36.0 & 36.1 & 42.0 & 33.3 & 33.1 & 45.4 & 50.7 & 37.0 & 34.1 & 35.9 & 24.4 & 25.4 & 35.7\\
      STCFormer~\cite{STCFormer} & 243 & \underline{29.5} & 33.2 & \underline{30.6} & 31.0 & 33.0 & 38.0 & 30.4 & 29.4 & 41.8 & 45.2 & 33.6 & 29.5 & 31.6 & \underline{21.3} & \underline{22.6} & \underline{32.0}\\
      STCFormer-L~\cite{STCFormer} & 243 & \textbf{29.3} & 33.0 & 30.7 & 30.6 & \underline{32.7} & 38.2 & \textbf{29.7} & \textbf{28.8} & 42.2 & \underline{45.0} & \underline{33.3} & \underline{29.4} & \underline{31.5} & \textbf{20.9} & \textbf{22.3} & \textbf{31.8}\\
      UPS~\cite{foo2023unified} & 243 & 30.3 & \textbf{32.2} & 30.8 & 33.1 & \textbf{31.1} & \textbf{35.2} & 30.3 & 32.1 & \textbf{39.4} & 49.6 & \textbf{32.9} & \textbf{29.2} & 33.9 & 21.6 & 24.5 & 32.5\\
      \dag~MotionBERT~\cite{motionbert} & 243 & 30.8 & 32.8 & 32.4 & 28.7 & 34.3 & 38.9 & \underline{30.1} & 30.0 & 42.5 & 49.7 & 36.0 & 30.8 & 31.7 & 22.0 & 23.0 & 32.9\\
      HDFormer~\cite{chen2023hdformer} & 96 & 29.6 & 33.8 & 31.7 & 31.3 & 33.7 & \underline{37.7} & 30.6 & 31.0 & 41.4 & 47.6 & 35.0 & 30.9 & 33.7 & 25.3 & 23.6 & 33.1\\
      HSTFormer~\cite{qian2023hstformer} & 81 & 31.1 & 33.7 & 33.0 & 33.2 & 33.6 & 38.8 & 31.9 & 31.5 & 43.7 & 46.3 & 35.7 & 31.5 & 33.1 & 24.2 & 24.5 & 33.7\\
        MotionAGFormer-L & 243 & 31.0 & \underline{32.6} & 31.0 & \textbf{27.9} & 34.0 & 38.7 & 31.5 & 30.0 & 41.4 & 45.4 & 34.8 & 30.8 & \textbf{31.3} & 22.8 & 23.2 & 32.5\\
      \rowcolor{gray!10} PoseMamba-S & 243 & 32.1 & 34.8 & 34.1 & 30.7 & 36.0 & 41.7 & 33.0 & 32.1 & 44.5 & 53.8 & 36.5 & 32.3 & 34.2 & 25.0 & 25.2   & 35.0\\
      \rowcolor{gray!10} PoseMamba-B & 243 & 32.3 & 34.0 & 33.2 & 30.2 & 34.9 & 40.6 & 32.0 & 31.0 & 44.5 & 53.0 & 36.3 & 31.3 & 33.5 & 23.8 & 24.6 & 34.3\\
      \rowcolor{gray!10} PoseMamba-L & 243 & 31.0 & \textbf{32.2} & 32.5 & \underline{28.5} & 33.1 & 38.6 & 30.5 &\underline{29.2} & \underline{41.3} & 47.8 & 34.9 & 30.2 & 31.7 & 22.7 & 23.5 & 32.5\\
      \hline
    \end{tabular}
}
\label{tab:human3.6m-comparison-action}
\end{table*}
In order to thoroughly assess the effectiveness of our methods for each specific action, we have presented a comparison in \Cref{tab:human3.6m-comparison-action} that highlights the MPJPE and P-MPJPE error rates of our approach in relation to various alternative methods employed on the Human3.6M dataset. Our findings indicate that our proposed methods consistently outperform the other alternatives in terms of P1 error for numerous actions, including but not limited to Direction, Discuss, Pose, Purchase, Sitting, Smoke, Wait, and Walk Two. Furthermore, for the remaining actions assessed in the dataset, we achieved the second-best performance results, demonstrating that our methods maintain a competitive edge across a wide range of activities while falling short of the top position in only a few instances.

\subsection{Per-joint Error Comparison}
\begin{figure}[!ht]
      \centering
      \includegraphics[width=1\linewidth]{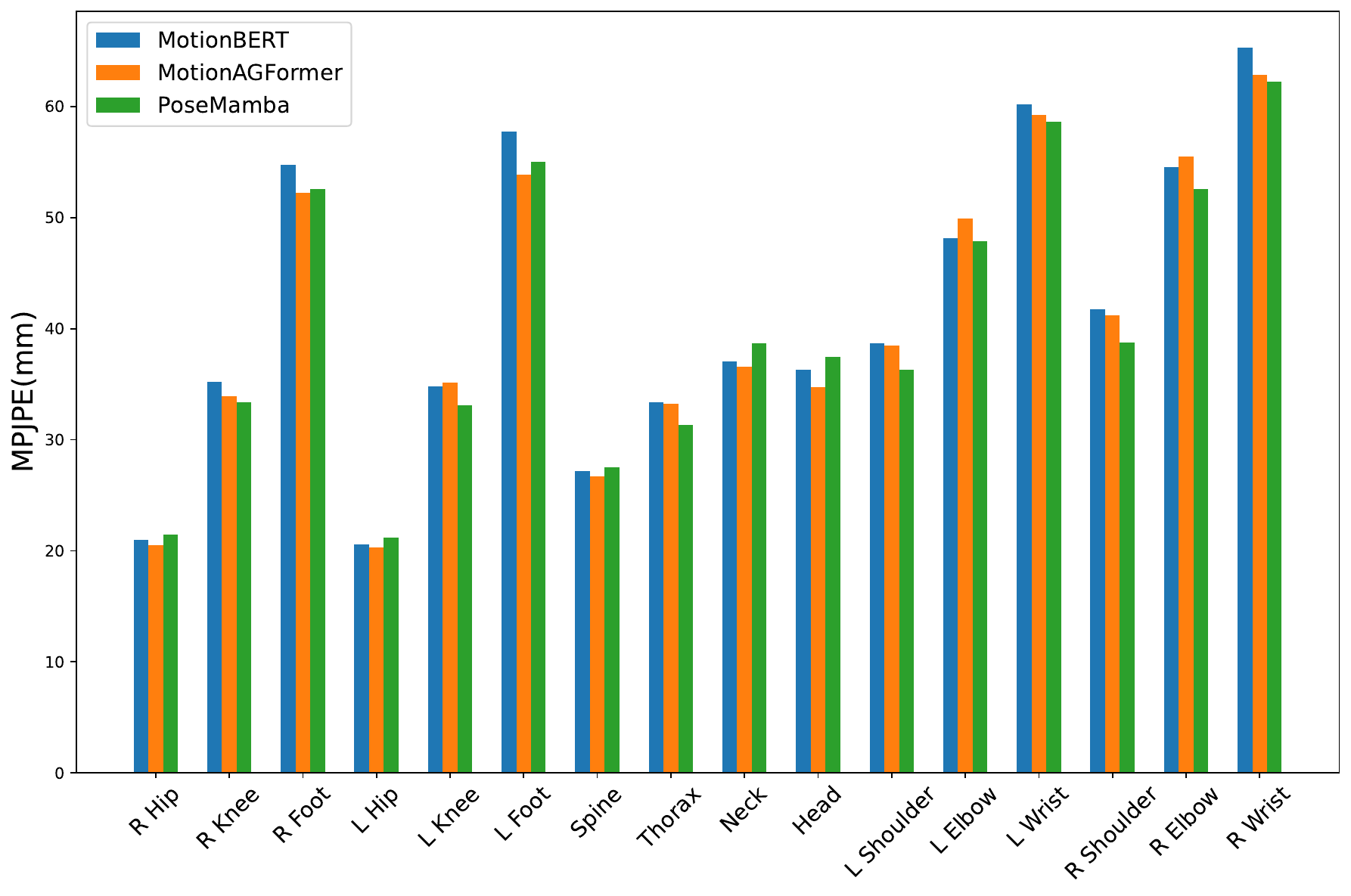}
      \caption{The per-joint error comparisons in terms of MPJPE (mm) with other models on Human3.6M dataset. 'L' and 'R' denote left and right, respectively.}
      \label{fig:per-joint-error}
\end{figure}

\Cref{fig:per-joint-error} provides a detailed comparison of the three models: MotionBERT, MotionAGFormer, and PoseMamba, specifically evaluated on the well-known Human3.6M benchmark dataset. In this analysis, it is observed that MotionAGFormer presents a slight competitive edge when it comes to the performance related to hip joint tracking, indicating that it may handle this specific part of the human body slightly better than the other models in question. However, this model experiences a notable disadvantage in terms of elbow joint performance when compared to its counterparts, which is worth mentioning. On the other hand, PoseMamba exhibits some challenges in accurately tracking movements of the head and neck regions, resulting in lower performance scores in those areas. Nevertheless, PoseMamba shines remarkably in assessing limb movements, particularly regarding the shoulders and elbows, where it demonstrates a considerable advantage over MotionBERT and MotionAGFormer. This pronounced proficiency in limb performance is extremely important, as it has a substantial effect on the overall error rates observed in the models. Such accuracy is especially critical in practical applications that demand precision, such as hand interaction scenarios and comprehensive gait analysis, where understanding limb dynamics can lead to better outcomes and insights.
\section{Additional visualization results}
\begin{figure*}[!ht]
\centering
\includegraphics[width=1\linewidth]{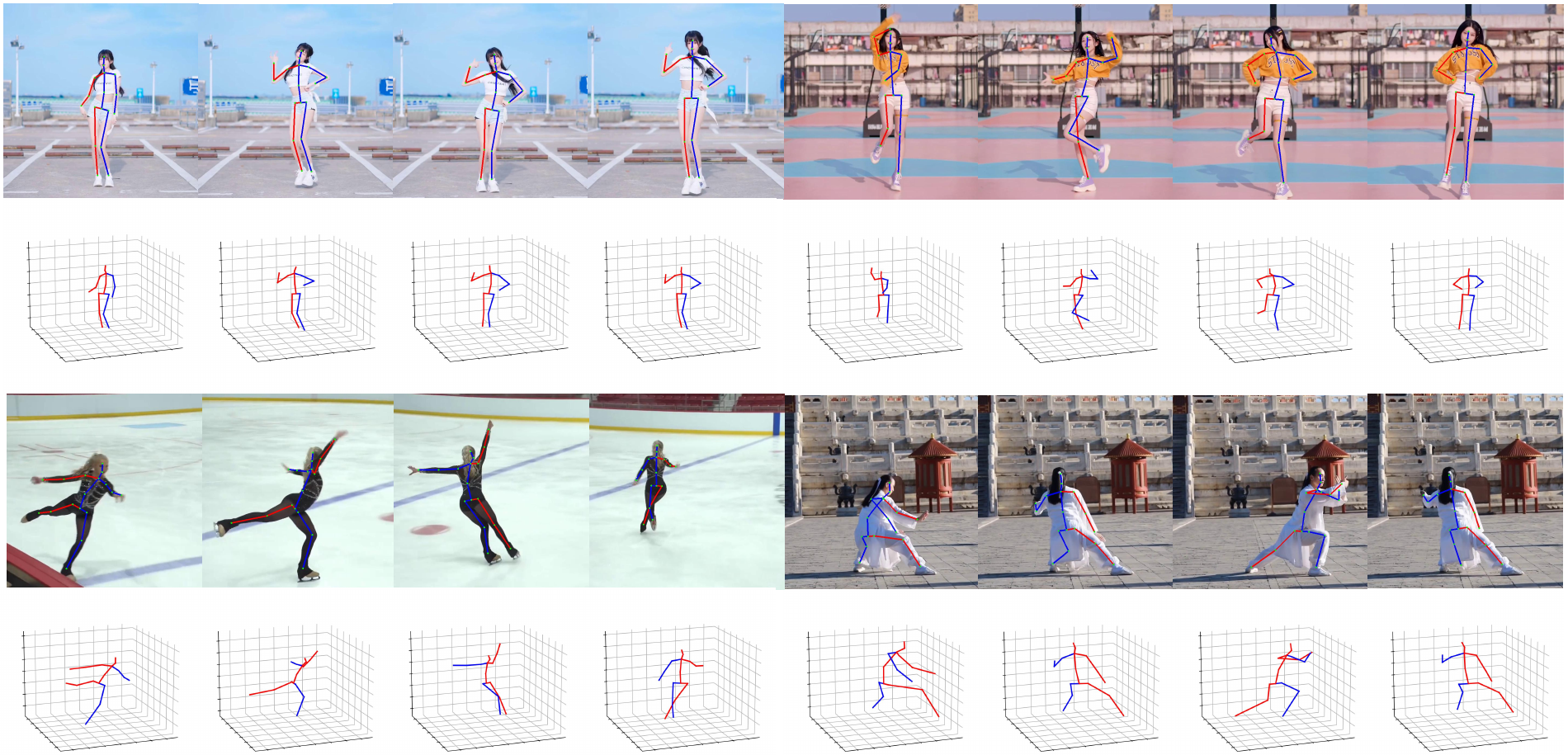}
\caption{Qualitative results of PoseMamba on challenging wild videos. Estimations are based on our large variant.}
\label{fig:in-the-wild}
\vspace{-0.3cm}
\end{figure*}

In order to verify the robustness and generalizability of PoseMamba, we conducted thorough testing using our large variant on several unseen videos captured in real-world scenarios. Our estimations indicate that even in challenging situations where the inferred 2D poses display some degree of noise—an instance that is particularly evident in the last frame of the fourth example illustrated in \Cref{fig:in-the-wild}—the resulting 3D pose estimations maintain a remarkable level of accuracy. This suggests that the model is quite resilient, effectively handling abrupt and noisy movements present within the 2D sequences. Furthermore, our results shows that when the subject is placed at a considerable distance from the camera, the 3D pose estimates still showcase a high level of precision and reliability, as demonstrated in the third example presented in \Cref{fig:in-the-wild}. This consistency reinforces the effectiveness of PoseMamba in various scenarios and distances, affirming its potential for practical applications.

\begin{figure*}[!ht]
\centering
\includegraphics[width=1\linewidth]{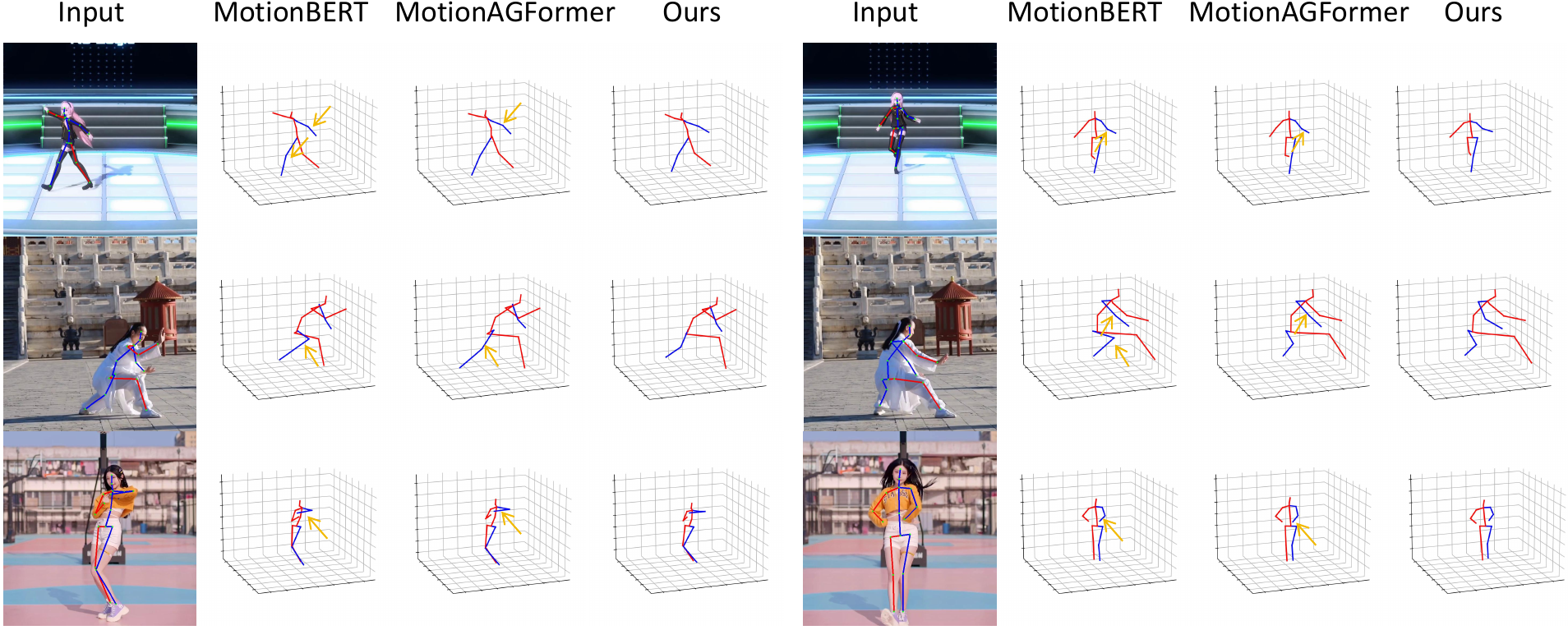}
\caption{Qualitative comparison among the proposed method (PoseMamba), the MotionBERT, and the MotionAGFormer on challenging wild videos. Wrong estimations are highlighted by yellow arrows. }
\label{fig:wild_compare}
\end{figure*}

Furthermore, as illustrated in \Cref{fig:wild_compare}, we present a qualitative comparison of our approach alongside the MotionBERT and MotionAGFormer methods, specifically focusing on several wild videos that showcase a variety of human movements. From \Cref{fig:wild_compare}, it becomes evident that our method is capable of generating 3D poses that are not only more precise but also more contextually appropriate. This improvement is particularly noticeable in scenarios where human actions are both intricate and infrequent, highlighting the strengths of our method in understanding and accurately representing complex behaviors.

\bibliography{aaai25}